\documentclass{article}

\usepackage{times}
\usepackage{graphicx} 
\usepackage{subfigure}

\usepackage{natbib}

\usepackage{algorithm}
\usepackage{algorithmic}
\usepackage[algo2e,linesnumbered,lined,boxed,vlined]{algorithm2e}

\usepackage{amsfonts}
\usepackage{amssymb}

\usepackage{hyperref}

\usepackage[accepted]{icml2016}

\usepackage{comment}
\usepackage{amsthm}
\usepackage{mathtools}
\usepackage{tabularx}
\usepackage{multirow}

\newtheorem{proposition}{Proposition}

\def\b{\ensuremath\boldsymbol}

\usepackage{lastpage}
\usepackage{fancyhdr}
\usepackage{url}
\icmltitlerunning{Eigenvalue and Generalized Eigenvalue Problems: Tutorial}

\begin{document}

\twocolumn[
\icmltitle{Eigenvalue and Generalized Eigenvalue Problems: Tutorial}

\icmlauthor{Benyamin Ghojogh}{bghojogh@uwaterloo.ca}
\icmladdress{Department of Electrical and Computer Engineering, 
\\Machine Learning Laboratory, University of Waterloo, Waterloo, ON, Canada}
\icmlauthor{Fakhri Karray}{karray@uwaterloo.ca}
\icmladdress{Department of Electrical and Computer Engineering, 
\\Centre for Pattern Analysis and Machine Intelligence, University of Waterloo, Waterloo, ON, Canada}
\icmlauthor{Mark Crowley}{mcrowley@uwaterloo.ca}
\icmladdress{Department of Electrical and Computer Engineering, 
\\Machine Learning Laboratory, University of Waterloo, Waterloo, ON, Canada}

\icmlkeywords{Tutorial, Locally Linear Embedding}

\vskip 0.3in
]

\begin{abstract}
This paper is a tutorial for eigenvalue and generalized eigenvalue problems. We first introduce eigenvalue problem, eigen-decomposition (spectral decomposition), and generalized eigenvalue problem. Then, we mention the optimization problems which yield to the eigenvalue and generalized eigenvalue problems. We also provide examples from machine learning, including principal component analysis, kernel supervised principal component analysis, and Fisher discriminant analysis, which result in eigenvalue and generalized eigenvalue problems. Finally, we introduce the solutions to both eigenvalue and generalized eigenvalue problems.
\end{abstract}

\section{Introduction}

Eigenvalue and generalized eigenvalue problems play important roles in different fields of science, including machine learning, physics, statistics, and mathematics. 
In eigenvalue problem, the eigenvectors of a matrix represent the most important and informative directions of that matrix. For example, if the matrix is a covariance matrix of data, the eigenvectors represent the directions of the spread or variance of data and the corresponding eigenvalues are the magnitude of the spread in these directions \cite{jolliffe2011principal}.
In generalized eigenvalue problem, these directions are impacted by another matrix. If the other matrix is the identity matrix, this impact is canceled and we will have the eigenvalue problem capturing the directions of the maximum spread.

In this paper, we introduce the eigenvalue problem and generalized eigenvalue problem and we introduce their solutions.
We also introduce the optimization problems which yield to the eigenvalue and generalized eigenvalue problems.
Some examples of these optimization problems in machine learning are also introduced for better illustration. The examples include principal component analysis, kernel supervised principal component analysis, and Fisher discriminant analysis.

\section{Introducing Eigenvalue and Generalized Eigenvalue Problems}

In this section, we introduce the eigenvalue problem and generalized eigenvalue problem.

\subsection{Eigenvalue Problem}

The eigenvalue problem \cite{wilkinson1965algebraic,golub2012matrix} of a symmetric matrix $\b{A} \in \mathbb{R}^{d \times d}$ is defined as:
\begin{align}\label{equation_EigenProblem}
\b{A} \b{\phi}_i = \lambda_i \b{\phi}_i, ~~~~~ \forall i \in \{1, \dots, d\},
\end{align}
and in matrix form, it is:
\begin{align}\label{equation_EigenProblem_matrixForm}
\b{A} \b{\Phi} = \b{\Phi} \b{\Lambda},
\end{align}
where the columns of $\mathbb{R}^{d \times d} \ni \b{\Phi} := [\b{\phi}_1, \dots, \b{\phi}_d]$ are the eigenvectors and diagonal elements of $\mathbb{R}^{d \times d} \ni \b{\Lambda} := \textbf{diag}([\lambda_1, \dots, \lambda_d]^\top)$ are the eigenvalues. 
Note that $\b{\phi}_i \in \mathbb{R}^d$ and $\lambda_i \in \mathbb{R}$.

Note that for eigenvalue problem, the matrix $\b{A}$ can be non-symmetric. If the matrix is symmetric, its eigenvectors are orthogonal/orthonormal and if it is non-symmetric, its eigenvectors are not orthogonal/orthonormal.

The Eq. (\ref{equation_EigenProblem_matrixForm}) can be restated as:
\begin{align}
\b{A} \b{\Phi} = \b{\Phi} \b{\Lambda} &\implies \b{A} \underbrace{\b{\Phi} \b{\Phi}^\top}_{\b{I}} = \b{\Phi} \b{\Lambda} \b{\Phi}^\top \nonumber \\
& \implies \b{A} = \b{\Phi} \b{\Lambda} \b{\Phi}^\top = \b{\Phi} \b{\Lambda} \b{\Phi}^{-1}, \label{equation_eigen_decomposition}
\end{align}
where $\b{\Phi}^\top = \b{\Phi}^{-1}$ because $\b{\Phi}$ is an orthogonal matrix. Moreover, note that we always have $\b{\Phi}^\top \b{\Phi} = \b{I}$ for orthogonal $\b{\Phi}$ but we only have $\b{\Phi} \b{\Phi}^\top = \b{I}$ if ``all'' the columns of the orthogonal $\b{\Phi}$ exist (it is not truncated, i.e., it is a square matrix).
The Eq. (\ref{equation_eigen_decomposition}) is referred to as ``eigenvalue decomposition'', ``eigen-decomposition'', or ``spectral decomposition''.
Note that the eigenvalue decomposition is completely related to the singular value decomposition. This relation is discussed in Appendix \ref{section_appendix_SVD}. 

\subsection{Generalized Eigenvalue Problem}

The generalized eigenvalue problem \cite{parlett1998symmetric,golub2012matrix} of two symmetric matrices $\b{A} \in \mathbb{R}^{d \times d}$ and $\b{B} \in \mathbb{R}^{d \times d}$ is defined as:
\begin{align}\label{equation_generalizedEigenProblem}
\b{A} \b{\phi}_i = \lambda_i \b{B} \b{\phi}_i, ~~~~~ \forall i \in \{1, \dots, d\},
\end{align}
and in matrix form, it is:
\begin{align}\label{equation_generalizedEigenProblem_matrixForm}
\b{A} \b{\Phi} = \b{B} \b{\Phi} \b{\Lambda},
\end{align}
where the columns of $\mathbb{R}^{d \times d} \ni \b{\Phi} := [\b{\phi}_1, \dots, \b{\phi}_d]$ are the eigenvectors and diagonal elements of $\mathbb{R}^{d \times d} \ni \b{\Lambda} := \textbf{diag}([\lambda_1, \dots, \lambda_d]^\top)$ are the eigenvalues. 
Note that $\b{\phi}_i \in \mathbb{R}^d$ and $\lambda_i \in \mathbb{R}$.

The generalized eigenvalue problem of Eq. (\ref{equation_generalizedEigenProblem}) or (\ref{equation_generalizedEigenProblem_matrixForm}) is denoted by $(\b{A}, \b{B})$. The $(\b{A}, \b{B})$ is called ``pair'' or ``pencil'' \cite{parlett1998symmetric}. The order in the pair matters. 
The $\b{\Phi}$ and $\b{\Lambda}$ are called the generalized eigenvectors and eigenvalues of $(\b{A}, \b{B})$. 
The $(\b{\Phi}, \b{\Lambda})$ or $(\b{\phi}_i, \lambda_i)$ is called the ``eigenpair'' of the pair $(\b{A}, \b{B})$ in the literature \cite{parlett1998symmetric}.

Comparing Eqs. (\ref{equation_EigenProblem}) and (\ref{equation_generalizedEigenProblem}) or Eqs. (\ref{equation_EigenProblem_matrixForm}) and (\ref{equation_generalizedEigenProblem_matrixForm}) shows that the eigenvalue problem is a special case of the generalized eigenvalue problem where $\b{B} = \b{I}$.

\section{Eigenvalue Optimization}

In this section, we introduce the optimization problems which yield to the eigenvalue problem. 

\subsection{Optimization Form 1}

Consider the following optimization problem with the variable $\b{\phi} \in \mathbb{R}^d$:
\begin{equation}\label{equation_EigenProblem_optimization}
\begin{aligned}
& \underset{\b{\phi}}{\text{maximize}}
& & \b{\phi}^\top \b{A}\, \b{\phi}, \\
& \text{subject to}
& & \b{\phi}^\top \b{\phi} = 1,
\end{aligned}
\end{equation}
where $\b{A} \in \mathbb{R}^{d \times d}$.
The Lagrangian \cite{boyd2004convex} for Eq. (\ref{equation_EigenProblem_optimization}) is:
\begin{align*}
\mathcal{L} = \b{\phi}^\top \b{A}\, \b{\phi} - \lambda\, (\b{\phi}^\top \b{\phi} - 1),
\end{align*}
where $\lambda \in \mathbb{R}$ is the Lagrange multiplier.
Equating the derivative of Lagrangian to zero gives us:
\begin{align*}
& \mathbb{R}^d \ni \frac{\partial \mathcal{L}}{\partial \b{\phi}} = 2\b{A} \b{\phi} - 2\lambda \b{\phi} \overset{\text{set}}{=} 0 \implies \b{A} \b{\phi} = \lambda \b{\phi},
\end{align*}
which is an eigenvalue problem for $\b{A}$ according to Eq. (\ref{equation_EigenProblem}). The $\b{\phi}$ is the eigenvector of $\b{A}$ and the $\lambda$ is the eigenvalue.

As the Eq. (\ref{equation_EigenProblem_optimization}) is a \textit{maximization} problem, the eigenvector is the one having the largest eigenvalue.
If the Eq. (\ref{equation_EigenProblem_optimization}) is a \textit{minimization} problem, the eigenvector is the one having the smallest eigenvalue.

It is noteworthy that in Eq. (\ref{equation_EigenProblem_optimization}) or the next optimization forms which will be discussed, the constraint can be set to any constant and it should not be necessarily $1$. The reason is that when we take derivative of the Lagrangian with respect to the optimization variable and set it to zero, the derivative of constant becomes zero regardless of its value. 

\subsection{Optimization Form 2}

Consider the following optimization problem with the variable $\b{\Phi} \in \mathbb{R}^{d \times d}$:
\begin{equation}\label{equation_EigenProblem_optimization_matrixForm}
\begin{aligned}
& \underset{\b{\Phi}}{\text{maximize}}
& & \textbf{tr}(\b{\Phi}^\top \b{A}\, \b{\Phi}), \\
& \text{subject to}
& & \b{\Phi}^\top \b{\Phi} = \b{I},
\end{aligned}
\end{equation}
where $\b{A} \in \mathbb{R}^{d \times d}$, the $\textbf{tr}(.)$ denotes the trace of matrix, and $\b{I}$ is the identity matrix.
Note that according to the properties of trace, the objective function can be any of these: $\textbf{tr}(\b{\Phi}^\top \b{A}\, \b{\Phi}) = \textbf{tr}(\b{\Phi}\b{\Phi}^\top \b{A}) = \textbf{tr}(\b{A}\b{\Phi}\b{\Phi}^\top)$.

The Lagrangian \cite{boyd2004convex} for Eq. (\ref{equation_EigenProblem_optimization_matrixForm}) is:
\begin{align}\label{equation_Lagrangian_with_Lambda}
\mathcal{L} = \textbf{tr}(\b{\Phi}^\top \b{A}\, \b{\Phi}) - \textbf{tr}\big(\b{\Lambda}^\top (\b{\Phi}^\top \b{\Phi} - \b{I})\big),
\end{align}
where $\b{\Lambda} \in \mathbb{R}^{d \times d}$ is a diagonal matrix whose entries are the Lagrange multipliers.
See Appendix \ref{section_proof_Lambda_diagonal} for proof of why the Lagrange multiplier matrix is diagonal. 

Equating derivative of $\mathcal{L}$ to zero gives us: 
\begin{align}
&\mathbb{R}^{d \times d} \ni \frac{\partial \mathcal{L}}{\partial \b{\Phi}} = 2\,\b{A} \b{\Phi} - 2\,\b{\Phi} \b{\Lambda} \overset{\text{set}}{=} \b{0} \nonumber \\
&\implies \b{A} \b{\Phi} = \b{\Phi} \b{\Lambda}, \label{equation_solution_optimization_form_2}
\end{align}
which is an eigenvalue problem for $\b{A}$ according to Eq. (\ref{equation_EigenProblem_matrixForm}).
The columns of $\b{\Phi}$ are the eigenvectors of $\b{A}$ and the diagonal elements of $\b{\Lambda}$ are the eigenvalues.

As the Eq. (\ref{equation_EigenProblem_optimization_matrixForm}) is a \textit{maximization} problem, the eigenvalues and eigenvectors in $\b{\Lambda}$ and $\b{\Phi}$ are sorted from the largest to smallest eigenvalues. 
If the Eq. (\ref{equation_EigenProblem_optimization_matrixForm}) is a \textit{minimization} problem, the eigenvalues and eigenvectors in $\b{\Lambda}$ and $\b{\Phi}$ are sorted from the smallest to largest eigenvalues (see Appendix \ref{section_sorting_eigenvalues}). 

\subsection{Optimization Form 3}

Consider the following optimization problem with the variable $\b{\phi} \in \mathbb{R}^{d}$:
\begin{equation}\label{equation_EigenProblem_optimization_reconstructionForm}
\begin{aligned}
& \underset{\b{\phi}}{\text{minimize}}
& & ||\b{X} - \b{\phi}\,\b{\phi}^\top\b{X}||_F^2, \\
& \text{subject to}
& & \b{\phi}^\top \b{\phi} = 1,
\end{aligned}
\end{equation}
where $\b{X} \in \mathbb{R}^{d \times n}$ and $||.||_F$ denotes the Frobenius norm of matrix.

The objective function in Eq. (\ref{equation_EigenProblem_optimization_reconstructionForm}) is simplified as:
\begin{align*}
||\b{X} -\, &\b{\phi}\b{\phi}^\top\b{X}||_F^2 \\
&= \textbf{tr}\big( (\b{X} - \b{\phi}\b{\phi}^\top\b{X})^\top (\b{X} - \b{\phi}\b{\phi}^\top\b{X}) \big) \\
&= \textbf{tr}\big( (\b{X}^\top - \b{X}^\top \b{\phi}\b{\phi}^\top) (\b{X} - \b{\phi}\b{\phi}^\top\b{X}) \big) \\
&= \textbf{tr}\big( \b{X}^\top\b{X} - 2\b{X}^\top\b{\phi}\b{\phi}^\top\b{X} + \b{X}^\top\b{\phi}\underbrace{\b{\phi}^\top\b{\phi}}_{1}\b{\phi}^\top\b{X} \big) \\
&= \textbf{tr}(\b{X}^\top\b{X}-\b{X}^\top\b{\phi}\b{\phi}^\top\b{X}) \\
&= \textbf{tr}(\b{X}^\top\b{X})-\textbf{tr}(\b{X}^\top\b{\phi}\b{\phi}^\top\b{X}) \\
&\overset{(a)}{=} \textbf{tr}(\b{X}^\top\b{X})-\textbf{tr}(\b{X}\b{X}^\top\b{\phi}\b{\phi}^\top) \\
&= \textbf{tr}(\b{X}^\top\b{X}-\b{X}\b{X}^\top\b{\phi}\b{\phi}^\top),
\end{align*}
where $(a)$ is because of the cyclic property of trace. 
The Lagrangian \cite{boyd2004convex} is: 
\begin{align*}
\mathcal{L} &= ||\b{X} - \b{\phi}\b{\phi}^\top\b{X}||_F^2 - \lambda (\b{\phi}^\top \b{\phi} - 1) \\
&= \textbf{tr}(\b{X}^\top\b{X})-\textbf{tr}(\b{X}\b{X}^\top\b{\phi}\b{\phi}^\top) - \lambda (\b{\phi}^\top \b{\phi} - 1),
\end{align*}
where $\lambda$ is the Lagrange multiplier. Equating the derivative of $\mathcal{L}$ to zero gives: 
\begin{align*}
&\mathbb{R}^d \ni \frac{\partial \mathcal{L}}{\partial \b{\phi}} = 2\,\b{X}\b{X}^\top \b{\phi} - 2\,\lambda\, \b{\phi} \overset{\text{set}}{=} \b{0} \\ 
&  \implies \b{X}\b{X}^\top \b{\phi} = \lambda\,\b{\phi} \overset{(a)}{\implies} \b{A}\, \b{\phi} = \lambda\,\b{\phi},
\end{align*}
where $(a)$ is because we take $\mathbb{R}^{d \times d} \ni \b{A} = \b{X}\b{X}^\top$.
The $\b{A}\, \b{\phi} = \lambda\,\b{\phi}$ is an eigenvalue problem for $\b{A}$ according to Eq. (\ref{equation_EigenProblem}). The $\b{\phi}$ is the eigenvector of $\b{A}$ and the $\lambda$ is the eigenvalue.

\subsection{Optimization Form 4}

Consider the following optimization problem with the variable $\b{\Phi} \in \mathbb{R}^{d \times d}$:
\begin{equation}\label{equation_EigenProblem_optimization_reconstructionForm_matrixFrom}
\begin{aligned}
& \underset{\b{\Phi}}{\text{minimize}}
& & ||\b{X} - \b{\Phi}\,\b{\Phi}^\top\b{X}||_F^2, \\
& \text{subject to}
& & \b{\Phi}^\top \b{\Phi} = \b{I},
\end{aligned}
\end{equation}
where $\b{X} \in \mathbb{R}^{d \times n}$.

Similar to what we had for Eq. (\ref{equation_EigenProblem_optimization_reconstructionForm}), the objective function in Eq. (\ref{equation_EigenProblem_optimization_reconstructionForm_matrixFrom}) is simplified as:
\begin{align*}
||\b{X} -\, &\b{\Phi}\b{\Phi}^\top\b{X}||_F^2 = \textbf{tr}(\b{X}^\top\b{X}-\b{X}\b{X}^\top\b{\Phi}\b{\Phi}^\top)
\end{align*}
The Lagrangian \cite{boyd2004convex} is: 
\begin{align*}
\mathcal{L} &= ||\b{X} - \b{\Phi}\b{\Phi}^\top\b{X}||_F^2 - \textbf{tr}\big(\b{\Lambda}^\top (\b{\Phi}^\top \b{\Phi} - \b{I})\big) \\
&= \textbf{tr}(\b{X}^\top\b{X})-\textbf{tr}(\b{X}\b{X}^\top\b{\Phi}\b{\Phi}^\top) \\
&~~~~~ - \textbf{tr}\big(\b{\Lambda}^\top (\b{\Phi}^\top \b{\Phi} - \b{I})\big),
\end{align*}
where $\b{\Lambda} \in \mathbb{R}^{d \times d}$ is a diagonal matrix (see Appendix \ref{section_proof_Lambda_diagonal}) including Lagrange multipliers. Equating the derivative of $\mathcal{L}$ to zero gives: 
\begin{align*}
&\mathbb{R}^{d \times d} \ni \frac{\partial \mathcal{L}}{\partial \b{\Phi}} = 2\,\b{X}\b{X}^\top \b{\Phi} - 2\,\b{\Phi} \b{\Lambda} \overset{\text{set}}{=} \b{0} \\
&\implies \b{X}\b{X}^\top \b{\Phi} = \b{\Phi} \b{\Lambda} \implies \b{A} \b{\Phi} = \b{\Phi} \b{\Lambda},
\end{align*}
which is an eigenvalue problem for $\b{A}$ according to Eq. (\ref{equation_EigenProblem_matrixForm}). The columns of $\b{\Phi}$ are the eigenvectors of $\b{A}$ and the diagonal elements of $\b{\Lambda}$ are the eigenvalues.

\subsection{Optimization Form 5}

Consider the following optimization problem with the variable $\b{\phi} \in \mathbb{R}^{d}$:
\begin{equation}\label{equation_EigenProblem_optimization_fractionalForm}
\begin{aligned}
& \underset{\b{\phi}}{\text{maximize}}
& & \frac{\b{\phi}^\top\, \b{A}\, \b{\phi}}{\b{\phi}^\top \b{\phi}}.
\end{aligned}
\end{equation}

According to the Rayleigh-Ritz quotient method \cite{croot2005rayleigh}, this optimization problem can be restated as (see Appendix \ref{section_appendix_rayleigh_ritz_quotient}):
\begin{equation}\label{equation_EigenProblem_optimization_fractionalForm2}
\begin{aligned}
& \underset{\b{\phi}}{\text{maximize}}
& & \b{\phi}^\top\, \b{A}\, \b{\phi}, \\
& \text{subject to}
& & \b{\phi}^\top \b{\phi} = 1,
\end{aligned}
\end{equation}
The Lagrangian \cite{boyd2004convex} is: 
\begin{align*}
\mathcal{L} = \b{\phi}^\top\, \b{A}\, \b{\phi} - \lambda (\b{\phi}^\top \b{\phi} - 1),
\end{align*}
where $\lambda$ is the Lagrange multiplier. Equating the derivative of $\mathcal{L}$ to zero gives: 
\begin{align*}
&\frac{\partial \mathcal{L}}{\partial \b{w}} = 2\,\b{A}\,\b{\phi} - 2\,\lambda\, \b{\phi} \overset{\text{set}}{=} \b{0} \\ 
&  \implies 2\,\b{A}\,\b{\phi} = 2\,\lambda\, \b{\phi} \implies \b{A}\,\b{\phi} = \lambda\, \b{\phi},
\end{align*}
which is an eigenvalue problem for $\b{A}$ according to Eq. (\ref{equation_EigenProblem}). The $\b{\phi}$ is the eigenvector of $\b{A}$ and the $\lambda$ is the eigenvalue.

As the Eq. (\ref{equation_EigenProblem_optimization_fractionalForm}) is a \textit{maximization} problem, the eigenvector is the one having the largest eigenvalue.
If the Eq. (\ref{equation_EigenProblem_optimization_fractionalForm}) is a \textit{minimization} problem, the eigenvector is the one having the smallest eigenvalue.

\section{Generalized Eigenvalue Optimization}

In this section, we introduce the optimization problems which yield to the generalized eigenvalue problem. 

\subsection{Optimization Form 1}

Consider the following optimization problem with the variable $\b{\phi} \in \mathbb{R}^d$:
\begin{equation}\label{equation_GeneralizedEigenProblem_optimization}
\begin{aligned}
& \underset{\b{\phi}}{\text{maximize}}
& & \b{\phi}^\top \b{A}\, \b{\phi}, \\
& \text{subject to}
& & \b{\phi}^\top \b{B}\, \b{\phi} = 1,
\end{aligned}
\end{equation}
where $\b{A} \in \mathbb{R}^{d \times d}$ and $\b{B} \in \mathbb{R}^{d \times d}$.
The Lagrangian \cite{boyd2004convex} for Eq. (\ref{equation_GeneralizedEigenProblem_optimization}) is:
\begin{align*}
\mathcal{L} = \b{\phi}^\top \b{A}\, \b{\phi} - \lambda\, (\b{\phi}^\top \b{B}\, \b{\phi} - 1),
\end{align*}
where $\lambda \in \mathbb{R}$ is the Lagrange multiplier.
Equating the derivative of Lagrangian to zero gives us:
\begin{align*}
& \mathbb{R}^d \ni \frac{\partial \mathcal{L}}{\partial \b{\phi}} = 2\b{A} \b{\phi} - 2\lambda \b{B} \b{\phi} \overset{\text{set}}{=} 0 \implies \b{A} \b{\phi} = \lambda \b{B}\b{\phi},
\end{align*}
which is a generalized eigenvalue problem $(\b{A}, \b{B})$ according to Eq. (\ref{equation_generalizedEigenProblem}). The $\b{\phi}$ is the eigenvector and the $\lambda$ is the eigenvalue for this problem.

As the Eq. (\ref{equation_GeneralizedEigenProblem_optimization}) is a \textit{maximization} problem, the eigenvector is the one having the largest eigenvalue.
If the Eq. (\ref{equation_GeneralizedEigenProblem_optimization}) is a \textit{minimization} problem, the eigenvector is the one having the smallest eigenvalue.

Comparing Eqs. (\ref{equation_EigenProblem_optimization}) and (\ref{equation_GeneralizedEigenProblem_optimization}) shows that eigenvalue problem is a special case of generalized eigenvalue problem where $\b{B} = \b{I}$.

\subsection{Optimization Form 2}

Consider the following optimization problem with the variable $\b{\Phi} \in \mathbb{R}^{d \times d}$:
\begin{equation}\label{equation_GeneralizedEigenProblem_optimization_matrixForm}
\begin{aligned}
& \underset{\b{\Phi}}{\text{maximize}}
& & \textbf{tr}(\b{\Phi}^\top \b{A}\, \b{\Phi}), \\
& \text{subject to}
& & \b{\Phi}^\top \b{B}\, \b{\Phi} = \b{I},
\end{aligned}
\end{equation}
where $\b{A} \in \mathbb{R}^{d \times d}$ and $\b{B} \in \mathbb{R}^{d \times d}$.
Note that according to the properties of trace, the objective function can be any of these: $\textbf{tr}(\b{\Phi}^\top \b{A}\, \b{\Phi}) = \textbf{tr}(\b{\Phi}\b{\Phi}^\top \b{A}) = \textbf{tr}(\b{A}\b{\Phi}\b{\Phi}^\top)$.

The Lagrangian \cite{boyd2004convex} for Eq. (\ref{equation_GeneralizedEigenProblem_optimization_matrixForm}) is:
\begin{align*}
\mathcal{L} = \textbf{tr}(\b{\Phi}^\top \b{A}\, \b{\Phi}) - \textbf{tr}\big(\b{\Lambda}^\top (\b{\Phi}^\top \b{B}\, \b{\Phi} - \b{I})\big),
\end{align*}
where $\b{\Lambda} \in \mathbb{R}^{d \times d}$ is a diagonal matrix (see Appendix \ref{section_proof_Lambda_diagonal}) whose entries are the Lagrange multipliers.

Equating derivative of $\mathcal{L}$ to zero gives us: 
\begin{align*}
&\mathbb{R}^{d \times d} \ni \frac{\partial \mathcal{L}}{\partial \b{\Phi}} = 2\,\b{A} \b{\Phi} - 2\,\b{B}\b{\Phi} \b{\Lambda} \overset{\text{set}}{=} \b{0} \\
&\implies \b{A} \b{\Phi} = \b{B}\b{\Phi} \b{\Lambda},
\end{align*}
which is an eigenvalue problem $(\b{A}, \b{B})$ according to Eq. (\ref{equation_generalizedEigenProblem_matrixForm}).
The columns of $\b{\Phi}$ are the eigenvectors of $\b{A}$ and the diagonal elements of $\b{\Lambda}$ are the eigenvalues.

As the Eq. (\ref{equation_GeneralizedEigenProblem_optimization_matrixForm}) is a \textit{maximization} problem, the eigenvalues and eigenvectors in $\b{\Lambda}$ and $\b{\Phi}$ are sorted from the largest to smallest eigenvalues. 
If the Eq. (\ref{equation_GeneralizedEigenProblem_optimization_matrixForm}) is a \textit{minimization} problem, the eigenvalues and eigenvectors in $\b{\Lambda}$ and $\b{\Phi}$ are sorted from the smallest to largest eigenvalues (see Appendix \ref{section_sorting_eigenvalues}). 

\subsection{Optimization Form 3}

Consider the following optimization problem \cite{parlett1998symmetric} with the variable $\b{\phi} \in \mathbb{R}^{d}$:
\begin{equation}\label{equation_GeneralizedEigenProblem_optimization_fractionalForm}
\begin{aligned}
& \underset{\b{\phi}}{\text{maximize}}
& & \frac{\b{\phi}^\top\, \b{A}\, \b{\phi}}{\b{\phi}^\top \b{B}\, \b{\phi}}.
\end{aligned}
\end{equation}

According to the generalized Rayleigh-Ritz quotient method \cite{croot2005rayleigh}, this optimization problem can be restated as (see Appendix \ref{section_appendix_rayleigh_ritz_quotient}):
\begin{equation}\label{equation_GeneralizedEigenProblem_optimization_fractionalForm2}
\begin{aligned}
& \underset{\b{\phi}}{\text{maximize}}
& & \b{\phi}^\top\, \b{A}\, \b{\phi}, \\
& \text{subject to}
& & \b{\phi}^\top \b{B}\, \b{\phi} = 1,
\end{aligned}
\end{equation}
The Lagrangian \cite{boyd2004convex} is: 
\begin{align*}
\mathcal{L} = \b{\phi}^\top\, \b{A}\, \b{\phi} - \lambda (\b{\phi}^\top \b{B}\, \b{\phi} - 1),
\end{align*}
where $\lambda$ is the Lagrange multiplier. Equating the derivative of $\mathcal{L}$ to zero gives: 
\begin{align*}
&\frac{\partial \mathcal{L}}{\partial \b{w}} = 2\,\b{A}\,\b{\phi} - 2\,\lambda\, \b{B}\, \b{\phi} \overset{\text{set}}{=} \b{0} \\ 
&  \implies 2\,\b{A}\,\b{\phi} = 2\,\lambda\, \b{B}\, \b{\phi} \implies \b{A}\,\b{\phi} = \lambda\, \b{B}\, \b{\phi},
\end{align*}
which is a generalized eigenvalue problem $(\b{A}, \b{B})$ according to Eq. (\ref{equation_generalizedEigenProblem}). 
The $\b{\phi}$ is the eigenvector and the $\lambda$ is the eigenvalue.

As the Eq. (\ref{equation_GeneralizedEigenProblem_optimization_fractionalForm}) is a \textit{maximization} problem, the eigenvector is the one having the largest eigenvalue.
If the Eq. (\ref{equation_GeneralizedEigenProblem_optimization_fractionalForm}) is a \textit{minimization} problem, the eigenvector is the one having the smallest eigenvalue.

\section{Examples for the Optimization Problems}

In this section, we introduce some examples in machine learning which use the introduced optimization problems.

\subsection{Examples for Eigenvalue Problem}

\subsubsection{Variance in Principal Component Analysis}

In Principal Component Analysis (PCA) \cite{pearson1901liii,friedman2001elements}, if we want to project onto one vector (one-dimensional PCA subspace), the problem is:
\begin{equation}
\begin{aligned}
& \underset{\b{u}}{\text{maximize}}
& & \b{u}^\top \b{S}\, \b{u}, \\
& \text{subject to}
& & \b{u}^\top \b{u} = 1,
\end{aligned}
\end{equation}
where $\b{u}$ is the projection direction and $\b{S}$ is the covariance matrix. Therefore, $\b{u}$ is the eigenvector of $\b{S}$ with the largest eigenvalue.

If we want to project onto a PCA subspace spanned by several directions, we have:
\begin{equation}
\begin{aligned}
& \underset{\b{U}}{\text{maximize}}
& & \textbf{tr}(\b{U}^\top \b{S}\, \b{U}), \\
& \text{subject to}
& & \b{U}^\top \b{U} = \b{I},
\end{aligned}
\end{equation}
where the columns of $\b{U}$ span the PCA subspace.

\subsubsection{Reconstruction in Principal Component Analysis}

We can look at PCA with another perspective: PCA is the best linear projection which has the smallest reconstruction error.
If we have one PCA direction, the projection is $\b{u}^\top \b{X}$ and the reconstruction is $\b{u}\b{u}^\top \b{X}$. We want the error between the reconstructed data and the original data to be minimized:
\begin{equation}
\begin{aligned}
& \underset{\b{u}}{\text{minimize}}
& & ||\b{X} - \b{u}\,\b{u}^\top\b{X}||_F^2, \\
& \text{subject to}
& & \b{u}^\top \b{u} = 1.
\end{aligned}
\end{equation}
Therefore, $\b{u}$ is the eigenvector of the covariance matrix $\b{S} = \b{X}\b{X}^\top$ (the $\b{X}$ is already centered by removing its mean).

If we consider several PCA directions, i.e., the columns of $\b{U}$, the minimization of the reconstruction error is:
\begin{equation}
\begin{aligned}
& \underset{\b{U}}{\text{minimize}}
& & ||\b{X} - \b{U}\,\b{U}^\top\b{X}||_F^2, \\
& \text{subject to}
& & \b{U}^\top \b{U} = \b{I}.
\end{aligned}
\end{equation}
Thus, the columns of $\b{U}$ are the eigenvectors of the covariance matrix $\b{S} = \b{X}\b{X}^\top$ (the $\b{X}$ is already centered by removing its mean).

\subsection{Examples for Generalized Eigenvalue Problem}

\subsubsection{Kernel Supervised Principal Component Analysis}

Kernel Supervised PCA (SPCA) \cite{barshan2011supervised} uses the following optimization problem:
\begin{equation}\label{equation_kernel_SPCA}
\begin{aligned}
& \underset{\b{\Theta}}{\text{maximize}}
& & \textbf{tr}(\b{\Theta}^\top \b{K}_x \b{H}\b{K}_y\b{H}\b{K}_x \b{\Theta}), \\
& \text{subject to}
& & \b{\Theta}^\top \b{K}_x\, \b{\Theta} = \b{I},
\end{aligned}
\end{equation}
where $\b{K}_x$ and $\b{K}_y$ are the kernel matrices over the training data and the labels of the training data, respectively, the $\b{H} := \b{I} - (1/n) \b{1}\b{1}^\top$ is the centering matrix, and the columns of $\b{\Theta}$ span the kernel SPCA subspace.

According to Eq. (\ref{equation_GeneralizedEigenProblem_optimization_matrixForm}), the solution to Eq. (\ref{equation_kernel_SPCA}) is:
\begin{align}
\b{K}_x \b{H}\b{K}_y\b{H}\b{K}_x \b{\Theta} = \b{K}_x\b{\Theta} \b{\Lambda}, 
\end{align}
which is the generalized eigenvalue problem $(\b{K}_x \b{H}\b{K}_y\b{H}\b{K}_x, \b{K}_x)$ according to Eq. (\ref{equation_generalizedEigenProblem_matrixForm}) where the $\b{\Theta}$ and $\b{\Lambda}$ are the eigenvector and eigenvalue matrices, respectively.

\subsubsection{Fisher Discriminant Analysis}

Another example is Fisher Discriminant Analysis (FDA) \cite{fisher1936use,friedman2001elements} in which the Fisher criterion \cite{xu2006analysis} is maximized:
\begin{equation}\label{equation_Fisher_optimization}
\begin{aligned}
& \underset{\b{w}}{\text{maximize}}
& & \frac{\b{w}^\top \b{S}_B\, \b{w}}{\b{w}^\top \b{S}_W\, \b{w}}, \\
\end{aligned}
\end{equation}
where $\b{w}$ is the projection direction and $\b{S}_B$ and $\b{S}_W$ are between- and within-class scatters:
\begin{align}
&\b{S}_B = \sum_{j=1}^c (\b{\mu}_i - \b{\mu}_t) (\b{\mu}_i - \b{\mu}_t)^\top, \\
&\b{S}_W = \sum_{j=1}^c \sum_{i=1}^{n_j} (\b{x}_{j,i} - \b{\mu}_i) (\b{x}_{j,i} - \b{\mu}_i)^\top,
\end{align}
$c$ is the number of classes, $n_j$ is the sample size of the $j$-th class, $\b{x}_{j,i}$ is the $i$-th data point in the $j$-th class, $\b{\mu}_i$ is the mean of the $i$-th class, and $\b{\mu}_t$ is the total mean.

According to Rayleigh-Ritz quotient method \cite{croot2005rayleigh}, the optimization problem in Eq. (\ref{equation_Fisher_optimization}) can be restated as:
\begin{equation}\label{equation_Fisher_optimization_2}
\begin{aligned}
& \underset{\b{w}}{\text{maximize}}
& & \b{w}^\top \b{S}_B\, \b{w}, \\
& \text{subject to}
& & \b{w}^\top \b{S}_W\, \b{w} = 1.
\end{aligned}
\end{equation}
The Lagrangian \cite{boyd2004convex} is: 
\begin{align*}
\mathcal{L} = \b{w}^\top \b{S}_B\, \b{w} - \lambda (\b{w}^\top \b{S}_W\, \b{w} - 1),
\end{align*}
where $\lambda$ is the Lagrange multiplier. Equating the derivative of $\mathcal{L}$ to zero gives: 
\begin{align*}
&\frac{\partial \mathcal{L}}{\partial \b{w}} = 2\,\b{S}_B\,\b{w} - 2\,\lambda\, \b{S}_W\, \b{w} \overset{\text{set}}{=} \b{0} \\ 
&  \implies 2\,\b{S}_B\,\b{w} = 2\,\lambda\, \b{S}_W\, \b{w} \implies \b{S}_B\,\b{w} = \lambda\, \b{S}_W\, \b{w},
\end{align*}
which is a generalized eigenvalue problem $(\b{S}_B, \b{S}_W)$ according to Eq. (\ref{equation_generalizedEigenProblem}). The $\b{w}$ is the eigenvector with the largest eigenvalue and the $\lambda$ is the corresponding eigenvalue.

\section{Solution to Eigenvalue Problem}

In this section, we introduce the solution to the eigenvalue problem.
Consider the Eq. (\ref{equation_EigenProblem}):
\begin{align}\label{eigenvalueProblem_solution}
\b{A} \b{\phi}_i = \lambda_i \b{\phi}_i \implies (\b{A} - \lambda_i\b{I})\, \b{\phi}_i = \b{0},
\end{align}
which is a linear system of equations.
According to Cramer's rule, a linear system of equations has non-trivial solutions if and only if the determinant vanishes. Therefore:
\begin{align}\label{eigenvalueProblem_solution_det}
\textbf{det}(\b{A} - \lambda_i\b{I}) = 0,
\end{align}
where $\textbf{det}(.)$ denotes the determinant of matrix.
The Eq. (\ref{eigenvalueProblem_solution_det}) gives us a $d$-degree polynomial equation which has $d$ roots (answers). Note that if the $\b{A}$ is not full rank (if it is a singular matrix), some of the roots will be zero. 
Moreover, if $\b{A}$ is positive semi-definite, i.e., $\b{A} \succeq 0$, all the roots are non-negative.

The roots (answers) from Eq. (\ref{eigenvalueProblem_solution_det}) are the eigenvalues of $\b{A}$. After finding the roots, we put every answer in Eq. (\ref{eigenvalueProblem_solution}) and find its corresponding eigenvector, $\b{\phi}_i \in \mathbb{R}^d$. Note that putting the root in Eq. (\ref{eigenvalueProblem_solution}) gives us a vector which can be normalized because the direction of the eigenvector is important and not its magnitude. The information of magnitude exists in its corresponding eigenvalue.

\section{Solution to Generalized Eigenvalue Problem}

In this section, we introduce the solution to the generalized eigenvalue problem. 
Recall the Eq. (\ref{equation_GeneralizedEigenProblem_optimization_fractionalForm}) again:
\begin{equation*}
\begin{aligned}
& \underset{\b{\phi}}{\text{maximize}}
& & \frac{\b{\phi}^\top\, \b{A}\, \b{\phi}}{\b{\phi}^\top \b{B}\, \b{\phi}}.
\end{aligned}
\end{equation*}
Let $\rho$ be this fraction named Rayleigh quotient \cite{croot2005rayleigh}:
\begin{align}
\rho(\b{u}; \b{A}, \b{B}) := \frac{\b{u}^\top\, \b{A}\, \b{u}}{\b{u}^\top \b{B}\, \b{u}}, ~~~~ \forall \b{u} \neq \b{0}.
\end{align}
The $\rho$ is stationary at $\b{\phi} \neq \b{0}$ if and only if:
\begin{align}\label{generalizedEigenvalueProblem_equation_system}
(\b{A} - \lambda\, \b{B})\, \b{\phi} = \b{0},
\end{align}
for some scalar $\lambda$ \cite{parlett1998symmetric}.
The Eq. (\ref{generalizedEigenvalueProblem_equation_system}) is a linear system of equations.
This system of equations can also be obtained from the Eq. (\ref{equation_generalizedEigenProblem}):
\begin{align}\label{generalizedEigenvalueProblem_equation_system_2}
\b{A} \b{\phi}_i = \lambda_i \b{B} \b{\phi}_i \implies (\b{A} - \lambda_i\b{B})\, \b{\phi}_i = \b{0}.
\end{align}

As we mentioned earlier, eigenvalue problem is a special case of generalized eigenvalue problem (where $\b{B} = \b{I}$) which is obvious by comparing Eqs. (\ref{eigenvalueProblem_solution}) and (\ref{generalizedEigenvalueProblem_equation_system_2}).

According to Cramer's rule, a linear system of equations has non-trivial solutions if and only if the determinant vanishes. Therefore:
\begin{align}\label{generalizedEigenvalueProblem_solution_det}
\textbf{det}(\b{A} - \lambda_i\, \b{B}) = 0.
\end{align}
Similar to the explanations for Eq. (\ref{eigenvalueProblem_solution_det}), we can solve for the roots of Eq. (\ref{generalizedEigenvalueProblem_solution_det}). However, note that the Eq. (\ref{generalizedEigenvalueProblem_solution_det}) is obtained from Eq. (\ref{equation_generalizedEigenProblem}) or (\ref{equation_GeneralizedEigenProblem_optimization_fractionalForm}) where only one eigenvector $\b{\phi}$ is considered.

For solving Eq. (\ref{equation_generalizedEigenProblem_matrixForm}) in general case, there exist two solutions for the generalized eigenvalue problem one of which is a quick and dirty solution and the other is a rigorous method. Both of the methods are explained in the following.

\subsection{The Quick \& Dirty Solution}

Consider the Eq. (\ref{equation_generalizedEigenProblem_matrixForm}) again:
\begin{align*}
\b{A} \b{\Phi} = \b{B} \b{\Phi} \b{\Lambda}.
\end{align*}
If $\b{B}$ is not singular (is invertible ), we can left-multiply the expressions by $\b{B}^{-1}$:
\begin{align}\label{GeneralizedEigenvalueProblem_solution}
\b{B}^{-1} \b{A} \b{\Phi} = \b{\Phi} \b{\Lambda} \overset{(a)}{\implies} \b{C} \b{\Phi} = \b{\Phi} \b{\Lambda},
\end{align}
where $(a)$ is because we take $\b{C} = \b{B}^{-1} \b{A}$.
The Eq. (\ref{GeneralizedEigenvalueProblem_solution}) is the eigenvalue problem for $\b{C}$ according to Eq. (\ref{equation_EigenProblem_matrixForm}) and can be solved using the approach of Eq. (\ref{eigenvalueProblem_solution_det}).

Note that even if $\b{B}$ is singular, we can use a numeric hack (which is a little dirty) and slightly strengthen its main diagonal in order to make it full rank:
\begin{align}
(\b{B} + \varepsilon \b{I})^{-1} \b{A} \b{\Phi} = \b{\Phi} \b{\Lambda} \implies \b{C} \b{\Phi} = \b{\Phi} \b{\Lambda},
\end{align}
where $\varepsilon$ is a very small positive number, e.g., $\varepsilon = 10^{-5}$, large enough to make $\b{B}$ full rank.

\subsection{The Rigorous Solution}


Consider the Eq. (\ref{equation_generalizedEigenProblem_matrixForm}) again:
\begin{align*}
\b{A} \b{\Phi} = \b{B} \b{\Phi} \b{\Lambda}.
\end{align*}
There exists a rigorous method to solve the generalized eigenvalue problem \cite{web_generalized_eigenproblem} which is explained in the following.

Consider the eigenvalue problem for $\b{B}$:
\begin{align}\label{equation_generalizedEigenProblem_eigendecomposition_B}
\b{B} \b{\Phi}_B = \b{\Phi}_B \b{\Lambda}_B,
\end{align}
where $\b{\Phi}_B$ and $\b{\Lambda}_B$ are the eigenvector and eigenvalue matrices of $\b{B}$, respectively.
Then, we have:
\begin{align}
\b{B} \b{\Phi}_B = \b{\Phi}_B \b{\Lambda}_B &\implies \b{\Phi}_B^{-1} \b{B} \b{\Phi}_B = \underbrace{\b{\Phi}_B^{-1} \b{\Phi}_B}_{\b{I}} \b{\Lambda}_B = \b{\Lambda}_B \nonumber \\
& \overset{(a)}{\implies} \b{\Phi}_B^\top \b{B} \b{\Phi}_B = \b{\Lambda}_B, \label{equation_generalizedEigenProblem_lambdaB}
\end{align}
where $(a)$ is because $\b{\Phi}_B$ is an orthogonal matrix (its columns are orthonormal) and thus $\b{\Phi}_B^{-1} = \b{\Phi}_B^\top$.

We multiply $\b{\Lambda}_B^{-1/2}$ to equation (\ref{equation_generalizedEigenProblem_lambdaB}) from left and right hand sides:
\begin{align*}
&\b{\Lambda}_B^{-1/2} \b{\Phi}_B^\top \b{B} \b{\Phi}_B \b{\Lambda}_B^{-1/2} = \b{\Lambda}_B^{-1/2} \b{\Lambda}_B \b{\Lambda}_B^{-1/2} = \b{I}, \\
&\implies \breve{\b{\Phi}}_B^\top \b{B} \breve{\b{\Phi}}_B = \b{I},
\end{align*}
where:
\begin{align}\label{equation_generalizedEigenProblem_phi_breve_B}
\breve{\b{\Phi}}_B := \b{\Phi}_B \b{\Lambda}_B^{-1/2}.
\end{align}
We define $\breve{\b{A}}$ as:
\begin{align}\label{equation_generalizedEigenProblem_breve_A}
\breve{\b{A}} := \breve{\b{\Phi}}_B^\top \b{A} \breve{\b{\Phi}}_B.
\end{align}
The $\breve{\b{A}}$ is symmetric because:
\begin{align*}
\breve{\b{A}}^\top = (\breve{\b{\Phi}}_B^\top \b{A} \breve{\b{\Phi}}_B)^\top \overset{(a)}{=} \breve{\b{\Phi}}_B^\top \b{A} \breve{\b{\Phi}}_B = \breve{\b{A}}.
\end{align*}
where $(a)$ notices that $\b{A}$ is symmetric.

The eigenvalue problem for $\breve{\b{A}}$ is:
\begin{align}\label{equation_generalizedEigenProblem_eigendecomposition_breve_A}
\breve{\b{A}} \b{\Phi}_A = \b{\Phi}_A \b{\Lambda}_A,
\end{align}
where $\b{\Phi}_A$ and $\b{\Lambda}_A$ are the eigenvector and eigenvalue matrices of $\breve{\b{A}}$.
Left-multiplying $\b{\Phi}_A^{-1}$ to equation (\ref{equation_generalizedEigenProblem_eigendecomposition_breve_A}) gives us:
\begin{align}
&\b{\Phi}_A^{-1} \breve{\b{A}} \b{\Phi}_A = \underbrace{\b{\Phi}_A^{-1} \b{\Phi}_A}_{\b{I}} \b{\Lambda}_A \overset{(a)}{\implies} \b{\Phi}_A^\top \breve{\b{A}} \b{\Phi}_A = \b{\Lambda}_A, \label{equation_generalizedEigenProblem_lambdaA_for_Phi_A}
\end{align}
where $(a)$ is because $\b{\Phi}_A$ is an orthogonal matrix (its columns are orthonormal), so $\b{\Phi}_A^{-1} = \b{\Phi}_A^\top$.
Note that $\b{\Phi}_A$ is an orthogonal matrix because $\breve{\b{A}}$ is symmetric (if the matrix is symmetric, its eigenvectors are orthogonal/orthonormal). 
The equation (\ref{equation_generalizedEigenProblem_lambdaA_for_Phi_A}) is diagonalizing the matrix $\breve{\b{A}}$.

Plugging equation (\ref{equation_generalizedEigenProblem_breve_A}) in equation (\ref{equation_generalizedEigenProblem_lambdaA_for_Phi_A}) gives us:
\begin{align}
&\b{\Phi}_A^\top \breve{\b{\Phi}}_B^\top \b{A} \breve{\b{\Phi}}_B \b{\Phi}_A = \b{\Lambda}_A \nonumber \\
&\overset{(\ref{equation_generalizedEigenProblem_phi_breve_B})}{\implies} \b{\Phi}_A^\top \b{\Lambda}_B^{-1/2} \b{\Phi}_B^\top \b{A} \b{\Phi}_B \b{\Lambda}_B^{-1/2} \b{\Phi}_A = \b{\Lambda}_A \nonumber \\
&\implies \b{\Phi}^\top \b{A} \b{\Phi} = \b{\Lambda}_A, \label{equation_generalizedEigenProblem_lambdaA_for_Phi}
\end{align}
where:
\begin{align}\label{equation_generalizedEigenProblem_Phi}
\b{\Phi} := \breve{\b{\Phi}}_B \b{\Phi}_A = \b{\Phi}_B \b{\Lambda}_B^{-1/2} \b{\Phi}_A.
\end{align}
The $\b{\Phi}$ also diagonalizes $\b{B}$ because ($\b{I}$ is a diagonal matrix):
\begin{align}
\b{\Phi}^\top \b{B} \b{\Phi} &\overset{(\ref{equation_generalizedEigenProblem_Phi})}{=} (\b{\Phi}_B \b{\Lambda}_B^{-1/2} \b{\Phi}_A)^\top \b{B} (\b{\Phi}_B \b{\Lambda}_B^{-1/2} \b{\Phi}_A) \nonumber \\
&= \b{\Phi}_A^\top \b{\Lambda}_B^{-1/2} (\b{\Phi}_B^\top \b{B} \b{\Phi}_B) \b{\Lambda}_B^{-1/2} \b{\Phi}_A \nonumber \\
&\overset{(\ref{equation_generalizedEigenProblem_lambdaB})}{=} \b{\Phi}_A^\top \underbrace{\b{\Lambda}_B^{-1/2} \b{\Lambda}_B \b{\Lambda}_B^{-1/2}}_{\b{I}} \b{\Phi}_A = \b{\Phi}_A^\top \b{\Phi}_A \nonumber \\
&\overset{(a)}{=} \b{\Phi}_A^{-1} \b{\Phi}_A = \b{I}, \label{equation_generalizedEigenProblem_B_diagonalized}
\end{align}
where $(a)$ is because $\b{\Phi}_A$ is an orthogonal matrix.
From equation (\ref{equation_generalizedEigenProblem_B_diagonalized}), we have:
\begin{align}
\b{\Phi}^\top \b{B} \b{\Phi} = \b{I} &\implies \b{\Phi}^\top \b{B} \b{\Phi} \b{\Lambda}_A = \b{\Lambda}_A \nonumber \\
& \overset{(\ref{equation_generalizedEigenProblem_lambdaA_for_Phi})}{\implies} \b{\Phi}^\top \b{B} \b{\Phi} \b{\Lambda}_A = \b{\Phi}^\top \b{A} \b{\Phi} \nonumber \\
&\overset{(a)}{\implies} \b{B} \b{\Phi} \b{\Lambda}_A = \b{A} \b{\Phi}, \label{equation_generalizedEigenProblem_with_Lambda_A}
\end{align}
where $(a)$ is because $\b{\Phi} \neq \b{0}$.

Comparing equations (\ref{equation_generalizedEigenProblem_matrixForm}) and (\ref{equation_generalizedEigenProblem_with_Lambda_A}) shows us:
\begin{align}\label{equation_generalizedEigenProblem_lambdaA_equal_Lambda}
\b{\Lambda}_A = \b{\Lambda}.
\end{align}

To summarize, for finding $\b{\Phi}$ and $\b{\Lambda}$ in Eq. (\ref{equation_generalizedEigenProblem_matrixForm}), we do the following steps (note that $\b{A}$ and $\b{B}$ are given):
\begin{enumerate}
\itemsep0em 
\item From Eq. (\ref{equation_generalizedEigenProblem_eigendecomposition_B}), we find $\b{\Phi}_B$ and $\b{\Lambda}_B$.
\item From Eq. (\ref{equation_generalizedEigenProblem_phi_breve_B}), we find $\breve{\b{\Phi}}_B$. In case $\b{\Lambda}_B^{1/2}$ is singular in Eq. (\ref{equation_generalizedEigenProblem_phi_breve_B}), we can use the numeric hack $\breve{\b{\Phi}}_B \approx \b{\Phi}_B (\b{\Lambda}_B^{1/2} + \varepsilon \b{I})^{-1}$ where $\varepsilon$ is a very small positive number, e.g., $\varepsilon = 10^{-5}$, large enough to make $\b{\Lambda}_B^{1/2}$ full rank.
\item From Eq. (\ref{equation_generalizedEigenProblem_breve_A}), we find $\breve{\b{A}}$.
\item From Eq. (\ref{equation_generalizedEigenProblem_eigendecomposition_breve_A}), we find $\b{\Phi}_A$ and $\b{\Lambda}_A$. From Eq. (\ref{equation_generalizedEigenProblem_lambdaA_equal_Lambda}), $\b{\Lambda}$ is found.
\item From Eq. (\ref{equation_generalizedEigenProblem_Phi}), we find $\b{\Phi}$.
\end{enumerate}
The above instructions are given as an algorithm in Algorithm \ref{algorithm_solution_generalizedEigenProblem}.

\SetAlCapSkip{0.5em}
\IncMargin{0.8em}
\begin{algorithm2e}[!t]
\DontPrintSemicolon
    $\b{\Phi}_B, \b{\Lambda}_B \gets \b{B} \b{\Phi}_B = \b{\Phi}_B \b{\Lambda}_B$\;
    $\breve{\b{\Phi}}_B \gets \breve{\b{\Phi}}_B = \b{\Phi}_B \b{\Lambda}_B^{-1/2} \approx \b{\Phi}_B (\b{\Lambda}_B^{1/2} + \varepsilon \b{I})^{-1}$\;
    $\breve{\b{A}} \gets \breve{\b{A}} = \breve{\b{\Phi}}_B^\top \b{A} \breve{\b{\Phi}}_B$\;
    $\b{\Phi}_A, \b{\Lambda}_A \gets \breve{\b{A}} \b{\Phi}_A = \b{\Phi}_A \b{\Lambda}_A$\;
    $\b{\Lambda} \gets \b{\Lambda} = \b{\Lambda}_A$\;
    $\b{\Phi} \gets \b{\Phi} = \breve{\b{\Phi}}_B \b{\Phi}_A$\;
    \textbf{return} $\b{\Phi}$ and $\b{\Lambda}$\;
\caption{Solution to the generalized eigenvalue problem $\b{A} \b{\Phi} = \b{B} \b{\Phi} \b{\Lambda}$.}\label{algorithm_solution_generalizedEigenProblem}
\end{algorithm2e}
\DecMargin{0.8em}

\section{Conclusion}

This paper was a tutorial paper introducing the eigenvalue and generalized eigenvalue problems. The problems were introduced, their optimization problems were mentioned, and some examples from machine learning were provided for them. Moreover, the solution to the eigenvalue and generalized eigenvalue problems were introduced.

\appendix

\section{Relation of Eigenvalue and Singular Value Decompositions}\label{section_appendix_SVD}

Consider a matrix $\b{A} \in \mathbb{R}^{\alpha \times \beta}$.
Singular Value Decomposition (SVD) \cite{stewart1993early} is one of the most well-known and effective matrix decomposition methods. It has two different forms, i.e., complete and incomplete.
There are different methods for obtaining this decomposition, one of which is Jordan's algorithm \cite{stewart1993early}. Here, we do not explain how to obtain SVD but we introduce different forms of SVD and their properties.

The ``complete SVD'' decomposes the matrix as:
\begin{align}
&\mathbb{R}^{\alpha \times \beta} \ni \b{A} = \b{U}\b{\Sigma}\b{V}^\top, \\
&\b{U} \in \mathbb{R}^{\alpha \times \alpha}, ~~ \b{V} \in \mathbb{R}^{\beta \times \beta}, ~~ \b{\Sigma} \in \mathbb{R}^{\alpha \times \beta}, \nonumber 
\end{align}
where the columns of $\b{U}$ and the columns of $\b{V}$ are called ``left singular vectors'' and ``right singular vectors'', respectively. 
In complete SVD, the $\b{\Sigma}$ is a \textit{rectangular} diagonal matrix whose main diagonal includes the ``singular values''. In cases $\alpha > \beta$ and $\alpha < \beta$, this matrix is in the forms:
\begin{align*}
\b{\Sigma} = 
\begin{bmatrix}
    \sigma_{1} & 0 & 0 \\
    \vdots & \ddots & \vdots \\
    0 & 0 & \sigma_{\beta} \\
    0 & 0 & 0 \\
    \vdots & \vdots & \vdots \\
    0 & 0 & 0
\end{bmatrix}
\text{and}
\begin{bmatrix}
    \sigma_{1} & 0 & 0 & 0 & \cdots & 0 \\
    \vdots & \ddots & \vdots & 0 & \cdots & 0 \\
    0 & 0 & \sigma_{\alpha} & 0 & \cdots & 0 \\
\end{bmatrix},
\end{align*}
respectively.
In other words, the number of singular values is $\min(\alpha, \beta)$.

The ``incomplete SVD'' decomposes the matrix as:
\begin{align}
&\mathbb{R}^{\alpha \times \beta} \ni \b{A} = \b{U}\b{\Sigma}\b{V}^\top, \\
&\b{U} \in \mathbb{R}^{\alpha \times k}, ~~ \b{V} \in \mathbb{R}^{\beta \times k}, ~~ \b{\Sigma} \in \mathbb{R}^{k \times k}, \nonumber 
\end{align}
where \cite{golub1970singular}:
\begin{align}
k := \min(\alpha, \beta),
\end{align}
and the columns of $\b{U}$ and the columns of $\b{V}$ are called ``left singular vectors'' and ``right singular vectors'', respectively. 
In incomplete SVD, the $\b{\Sigma}$ is a \textit{square} diagonal matrix whose main diagonal includes the ``singular values''. The matrix $\b{\Sigma}$ is in the form:
\begin{align*}
\b{\Sigma} = 
\begin{bmatrix}
    \sigma_{1} & 0 & 0 \\
    \vdots & \ddots & \vdots \\
    0 & 0 & \sigma_{k} 
\end{bmatrix}.
\end{align*}

Note that in both complete and incomplete SVD, the left singular vectors are orthonormal and the right singular vectors are also orthonormal; therefore, $\b{U}$ and $\b{V}$ are both orthogonal matrices so:
\begin{align}
\b{U}^\top \b{U} = \b{I}, \\
\b{V}^\top \b{V} = \b{I}.
\end{align}
If these orthogonal matrices are not truncated and thus are square matrices, e.g., for complete SVD, we also have:
\begin{align}
\b{U} \b{U}^\top = \b{I}, \\
\b{V} \b{V}^\top = \b{I}.
\end{align}

\begin{proposition}\label{proposition_SVD}
In both complete and incomplete SVD of matrix $\b{A}$, the left and right singular vectors are the eigenvectors of $\b{A}\b{A}^\top$ and $\b{A}^\top \b{A}$, respectively, and the singular values are the square root of eigenvalues of either $\b{A}\b{A}^\top$ or $\b{A}^\top \b{A}$.
\end{proposition}

\begin{proof}
We have:
\begin{align*}
\b{A} \b{A}^\top &= (\b{U}\b{\Sigma}\b{V}^\top) (\b{U}\b{\Sigma}\b{V}^\top)^\top = \b{U}\b{\Sigma}\underbrace{\b{V}^\top \b{V}}_{\b{I}} \b{\Sigma}\b{U}^\top \\
&= \b{U}\b{\Sigma} \b{\Sigma}\b{U}^\top = \b{U}\b{\Sigma}^2\b{U}^\top,
\end{align*}
which is eigen-decomposition of $\b{A} \b{A}^\top$ where the columns of $\b{U}$ are the eigenvectors and the diagonal of $\b{\Sigma}^2$ are the eigenvalues (see Eq. (\ref{equation_eigen_decomposition})) so the diagonal of $\b{\Sigma}$ are the square root of eigenvalues.
We also have:
\begin{align*}
\b{A}^\top \b{A} &= (\b{U}\b{\Sigma}\b{V}^\top)^\top (\b{U}\b{\Sigma}\b{V}^\top) = \b{V}\b{\Sigma}\underbrace{\b{U}^\top \b{U}}_{\b{I}}\b{\Sigma}\b{V}^\top \\
&= \b{V}\b{\Sigma}\b{\Sigma}\b{V}^\top = \b{V}\b{\Sigma}^2\b{V}^\top,
\end{align*}
which is the eigenvalue decomposition of $\b{A}^\top \b{A}$ where the columns of $\b{V}$ are the eigenvectors and the diagonal of $\b{\Sigma}^2$ are the eigenvalues so the diagonal of $\b{\Sigma}$ are the square root of eigenvalues. Q.E.D.
\end{proof}

\section{Proof of Why the Lagrange Multiplier is Diagonal in the Eigenvalue Problem}\label{section_proof_Lambda_diagonal}

In both eigenvalue and generalized eigenvalue problems, the Lagrange multiplier matrix is diagonal. 
For example, consider Eq. (\ref{equation_Lagrangian_with_Lambda}), i.e. $\mathcal{L} = \textbf{tr}(\b{\Phi}^\top \b{A}\, \b{\Phi}) - \textbf{tr}\big(\b{\Lambda}^\top (\b{\Phi}^\top \b{\Phi} - \b{I})\big)$, in which $\b{\Lambda}$ is a diagonal matrix. 
Here, we prove and discuss why this Lagrange multiplier matrix is diagonal. A similar proof can be provided for the Lagrange multiplier in the generalized eigenvalue problem. 

\subsection{Discussion on the Number of Constraints}\label{section_discussion_number_of_constraints}

First note that the orthogonality constraint $\b{\Phi}^\top \b{\Phi} = \b{I}$ embeds $d \times d$ constraints but, as some of them are symmetric, we only have $d(d-1)/2 + d$ constraints. In other words, $d$ constraints are for the $d$ columns of $\b{\Phi}$ to have unit length and $d(d-1)/2$ constraints are for every two columns of $\b{\Phi}$ to be orthogonal. Let $\{\lambda_1, \dots, \lambda_d\}$ be the Lagrange multipliers for columns of $\b{\Phi}$ to have unit length and let $\{\lambda^1, \dots, \lambda^{d(d-1)/2}\}$ be the Lagrange multipliers for the columns of $\b{\Phi}$ to be orthogonal. Then, the Lagrangian is:
\begin{align*}
\mathcal{L} = &\,\textbf{tr}(\b{\Phi}^\top \b{A}\, \b{\Phi}) - \lambda_1 (\b{\phi}_1^\top \b{\phi}_1 - 1) - \dots - \lambda_d (\b{\phi}_d^\top \b{\phi}_d - 1) \\
& - \lambda^1 (\b{\phi}_1^\top \b{\phi}_2 - 0) - \lambda^1 (\b{\phi}_2^\top \b{\phi}_1 - 0) - \dots \\
&- \lambda^{d(d-1)/2} (\b{\phi}_{d-1}^\top \b{\phi}_d - 0) - \lambda^{d(d-1)/2} (\b{\phi}_d^\top \b{\phi}_{d-1} - 0),
\end{align*}
where each $\lambda^j$ is repeated twice because of the symmetry of inner product between two columns. 
We can gather the terms having the dual variables together using trace to obtain Eq. (\ref{equation_Lagrangian_with_Lambda}). Setting derivative of this equation to zero gives Eq. (\ref{equation_solution_optimization_form_2}) which is in the form of an eigenvalue problem (see Eq. (\ref{equation_EigenProblem_matrixForm})). Therefore, one of the possible solutions is the eigenvalue problem of $\b{\Phi}$ in which the eigenvalue matrix $\b{\Lambda}$ is diagonal; hence, in one of the solutions, the Lagrange multiplier matrix must be diagonal. 

In fact, this problem can have many solutions\footnote{We thank Sebastian Ciobanu for helping in this part.}. 
If $\b{\Phi}$ is the optimal solution for Eq. (\ref{equation_EigenProblem_optimization_matrixForm}), any matrix $\b{M} := \b{\Phi} \b{V}$ is also a solution if $\b{V}$ is an orthogonal matrix. 
Consider Eq. (\ref{equation_solution_optimization_form_2}). As the matrix $\b{\Lambda}$ is symmetric, we can have its own eigenvalue decomposition as (see Eq. (\ref{equation_eigen_decomposition})):
\begin{align}\label{equation_Lambda_eigen_decomposition}
\b{\Lambda} = \b{V} \b{\Omega} \b{V}^\top,
\end{align}
where $\b{V}$ and $\b{\Omega}$ are the eigenvectors and eigenvalues of $\b{\Lambda}$. 
The eigenvectors are orthonormal so the matrix $\b{V}$ is a non-truncated orthogonal matrix; hence:
\begin{align}\label{equation_V_eigenvectros_orthogonal}
\b{V}^\top \b{V} = \b{V} \b{V}^\top = \b{I}.
\end{align}
Hence, Eq. (\ref{equation_solution_optimization_form_2}) can be stated as:
\begin{align}
&\b{A} \b{\Phi} = \b{\Phi} \b{\Lambda} \overset{(\ref{equation_Lambda_eigen_decomposition})}{=} \b{\Phi} \b{V} \b{\Omega} \b{V}^\top \overset{(a)}{\implies} \nonumber\\
&\b{A} \b{\Phi} \b{V} = \b{\Phi} \b{V} \b{\Omega} \underbrace{\b{V}^\top \b{V}}_{=\b{I}} \overset{(\ref{equation_V_eigenvectros_orthogonal})}{\implies} \b{A} \b{\Phi} \b{V} = \b{\Phi} \b{V} \b{\Omega}, \label{equation_solution_optimization_form_2_scaled}
\end{align}
where $(a)$ is because of right multiplication by the matrix $\b{V}$.
Eq. (\ref{equation_solution_optimization_form_2_scaled}) is an eigenvalue problem for $\b{A}$ and shows that $\b{\Phi}\b{V}$ is also a solution for this eigenvalue problem. 

\subsection{Proof of Being Diagonal}

Now, we prove that why the Lagrange multiplier in the Lagrangian, i.e. Eq. (\ref{equation_Lagrangian_with_Lambda}), a diagonal matrix\footnote{We thank Ali Saheb Pasand for helping in this part.}. 
Assume that $\b{\Lambda}$ is not necessarily diagonal but we know it is symmetric because of the discussions in Section \ref{section_discussion_number_of_constraints}.
We can diagonalize this symmetric matrix orthogonally using its eigenvalue decomposition as we have done in Eq. (\ref{equation_Lambda_eigen_decomposition}).
We can restate the Lagrangian (Eq. (\ref{equation_Lagrangian_with_Lambda})) as:
\begin{align}
\mathcal{L} = &\,\textbf{tr}(\b{\Phi}^\top \b{A}\, \b{\Phi}) - \textbf{tr}\big(\b{\Lambda}^\top (\b{\Phi}^\top \b{\Phi} - \b{I})\big) \nonumber\\
&\overset{(\ref{equation_Lambda_eigen_decomposition})}{=} \textbf{tr}(\b{\Phi}^\top \b{A}\, \b{\Phi}) - \textbf{tr}\big(\b{V} \b{\Omega} \b{V}^\top (\b{\Phi}^\top \b{\Phi} - \b{I})\big) \nonumber\\
&\overset{(a)}{=} \textbf{tr}(\b{\Phi}^\top \b{A}\, \b{\Phi}) - \textbf{tr}\big(\b{\Omega} \b{V}^\top (\b{\Phi}^\top \b{\Phi} - \b{I}) \b{V}\big) \nonumber\\
&= \textbf{tr}(\b{\Phi}^\top \b{A}\, \b{\Phi}) - \textbf{tr}\big(\b{\Omega} (\b{V}^\top \b{\Phi}^\top \b{\Phi} \b{V} - \b{V}^\top \b{V}) \big) \nonumber\\
&\overset{(b)}{=} \textbf{tr}(\b{\Phi}^\top \b{A}\, \b{\Phi}) - \textbf{tr}\big(\b{\Omega} (\b{M}^\top \b{M} - \b{I}) \big) \nonumber\\
&\overset{(c)}{=} \textbf{tr}(\b{\Phi}^\top \b{A}\, \b{\Phi}) - \textbf{tr}\big(\b{\Omega}^\top (\b{M}^\top \b{M} - \b{I}) \big), \label{equation_Lagrangian_inside_proof}
\end{align}
where $(a)$ is because of the cyclic property of trace, $(b)$ is because of Eq. (\ref{equation_V_eigenvectros_orthogonal}) and our definition $\b{M} := \b{\Phi} \b{V}$, and $(c)$ is because the matrix $\b{\Omega}$ is diagonal (see Eq. (\ref{equation_Lambda_eigen_decomposition})) so it is equal to its transpose. 

We have:
\begin{align*}
&\b{M} = \b{\Phi} \b{V} \implies \b{M} \b{V}^\top = \b{\Phi} \b{V} \b{V}^\top \overset{(\ref{equation_V_eigenvectros_orthogonal})}{=} \b{\Phi} \\
&\implies \b{\Phi}^\top = \b{V} \b{M}^\top.
\end{align*}
Therefore, the first term in Eq. (\ref{equation_Lagrangian_inside_proof}) can be restated as:
\begin{align*}
\textbf{tr}(\b{\Phi}^\top \b{A} \b{\Phi}) &= \textbf{tr}(\b{V} \b{M}^\top \b{A} \b{M} \b{V}^\top) \\
&\overset{(a)}{=} \textbf{tr}(\b{V}^\top \b{V} \b{M}^\top \b{A} \b{M}) \overset{(\ref{equation_V_eigenvectros_orthogonal})}{=} \textbf{tr}(\b{M}^\top \b{A} \b{M}).
\end{align*}
Hence, Eq. (\ref{equation_Lagrangian_inside_proof}), which is the Lagrangian, can be stated as:
\begin{align}
\mathcal{L} = \textbf{tr}(\b{M}^\top \b{A} \b{M}) - \textbf{tr}\big(\b{\Omega}^\top (\b{M}^\top \b{M} - \b{I}) \big), \label{equation_Lagrangian_inside_proof_2}
\end{align}
Comparing Eqs. (\ref{equation_Lagrangian_with_Lambda}) and (\ref{equation_Lagrangian_inside_proof_2}) shows that we can have a change of variable $\b{M} = \b{\Phi} \b{V}$ and state the Lagrange multiplier as a diagonal matrix (because $\b{\Omega}$ is diagonal in Eq. (\ref{equation_Lagrangian_inside_proof_2})). As both $\b{\Lambda}$ and $\b{\Omega}$ are dummy variables and can have any name, we can initially assume that $\b{\Lambda}$ is diagonal.

\section{Discussion on the Sorting of Eigenvectors and Eigenvalues}\label{section_sorting_eigenvalues}

We know that the second derivative shows the curvature direction of a function. 
Consider Eq. (\ref{equation_EigenProblem_optimization_matrixForm}) with the Lagrangian in Eq. (\ref{equation_Lagrangian_with_Lambda}). 
The second derivative of the Lagrangian is:
\begin{align*}
&\mathbb{R}^{d \times d} \ni \frac{\partial^2 \mathcal{L}}{\partial \b{\Phi}^2} = 2\,\b{A}^\top - 2\,\b{\Lambda}^\top = 2 (\b{A} - \b{\Lambda})^\top \overset{\text{set}}{=} \b{0} \\
&\implies \b{A} = \b{\Lambda}.
\end{align*}
Hence, for the second derivative, the matrix $\b{A}$ is equal to the eigenvalues $\b{\Lambda}$. 
The matrix $\b{A}$ is in the objective function of Eq. (\ref{equation_EigenProblem_optimization_matrixForm}). 
Therefore, if the optimization problem (\ref{equation_EigenProblem_optimization_matrixForm}) is maximization (resp. minimization), the eigenvalues, and their corresponding eigenvectors, should be sorted from largest to smallest (resp. from smallest to largest). 
Likewise, Consider Eq. (\ref{equation_GeneralizedEigenProblem_optimization_matrixForm}) for generalized eigenvalue problem. 
The second derivative of the Lagrangian is:
\begin{align*}
&\mathbb{R}^{d \times d} \ni \frac{\partial^2 \mathcal{L}}{\partial \b{\Phi}^2} = 2\,\b{A}^\top - 2\,\b{B}^\top \b{\Lambda}^\top = 2 (\b{A} - \b{\Lambda} \b{B})^\top \overset{\text{set}}{=} \b{0} \\
&\implies \b{A} = \b{\Lambda} \b{B}.
\end{align*}
Hence, for the second derivative, the matrix $\b{A}$ is related the eigenvalues $\b{\Lambda}$ and a similar analysis holds. We can also have a similar analysis for other forms of optimization for eigenvalue and generalized eigenvalue problems.

\section{Rayleigh-Ritz Quotient}\label{section_appendix_rayleigh_ritz_quotient}

\subsection{Rayleigh-Ritz Quotient}

The \textit{Rayleigh-Ritz quotient} or \textit{Rayleigh quotient} is defined as \cite{parlett1998symmetric,croot2005rayleigh}:
\begin{align}\label{equation_rayleigh_ritz_quotient}
\mathbb{R} \ni R(\b{A}, \b{x}) := \frac{\b{x}^\top \b{A}\, \b{x}}{\b{x}^\top \b{x}},
\end{align}
where $\b{A}$ is a symmetric matrix and $\b{x}$ is a non-zero vector:
\begin{align}
\b{A} = \b{A}^\top, ~~ \b{x} \neq \b{0}.
\end{align}
One of the properties of the Rayleigh-Ritz quotient is:
\begin{align}\label{equation_rayleigh_ritz_quotient_scaling}
R(\b{A}, c\b{x}) = R(\b{A}, \b{x}),
\end{align}
where $c$ is a scalar. The proof is that:
\begin{align*}
R(\b{A}, c\b{x}) &= \frac{(c\b{x})^\top \b{A}\, c\b{x}}{(c\b{x})^\top c\b{x}} \overset{(a)}{=} \frac{c\b{x}^\top \b{A}\, c\b{x}}{c\b{x}^\top c\b{x}} \\
&\overset{(b)}{=} \frac{c^2}{c^2} \times \frac{\b{x}^\top \b{A}\, \b{x}}{\b{x}^\top \b{x}} \overset{(\ref{equation_rayleigh_ritz_quotient})}{=} R(\b{A}, \b{x}),
\end{align*}
where $(a)$ and $(b)$ are because $c$ is a scalar.

Because of the Eq. (\ref{equation_rayleigh_ritz_quotient_scaling}), the optimization of the Rayleigh-Ritz quotient has an equivalent \cite{croot2005rayleigh}:
\begin{align}
&\underset{\b{x}}{\text{minimize/maximize}} ~~ R(\b{A}, \b{x}) \overset{(a)}{\equiv} \nonumber \\
&\begin{aligned}
& \underset{\b{x}}{\text{minimize/maximize}}
& & R(\b{A}, \b{x}) \\
& \text{subject to}
& & ||\b{x}||_2 = 1,
\end{aligned} \overset{(b)}{\equiv} \nonumber \\
&\begin{aligned}\label{equation_rayleigh_ritz_quotient_optimization}
& \underset{\b{x}}{\text{minimize/maximize}}
& & \b{x}^\top \b{A}\, \b{x} \\
& \text{subject to}
& & ||\b{x}||_2 = 1,
\end{aligned}
\end{align}
where $(a)$ is because if we define $\b{y} := (1/||\b{x}||_2)\, \b{x}$, the Rayleigh-Ritz quotient is:
\begin{align}
R(\b{A}, \b{y}) = \frac{\b{y}^\top \b{A}\, \b{y}}{\b{y}^\top \b{y}} = \frac{1/||\b{x}||_2^2}{1/||\b{x}||_2^2} \times \frac{\b{x}^\top \b{A}\, \b{x}}{\b{x}^\top \b{x}} = R(\b{A}, \b{x}),
\end{align}
and:
\begin{align}
||\b{y}||_2^2 = \frac{1}{||\b{x}||_2^2} \times ||\b{x}||_2^2 = 1 \implies ||\b{y}||_2 = 1.
\end{align}
Thus, we have $R(\b{A}, \b{y})$ subject to $||\b{y}||_2 = 1$. Changing the dummy variable $\b{y}$ to $\b{x}$ gives the Eq. (\ref{equation_rayleigh_ritz_quotient_optimization}).
The $(b)$ notices $\b{x}^\top\b{x} = 1$ because of the constraint $||\b{x}||_2=1$.

Note that the constraint in Eq. (\ref{equation_rayleigh_ritz_quotient_optimization}) can be equal to any constant which is proved similarly. Moreover, note that the value of constant in the constraint is not important because it will be removed after taking derivative from the Lagrangian in optimization \cite{boyd2004convex}.

\subsection{Generalized Rayleigh-Ritz Quotient}

The \textit{generalized Rayleigh-Ritz quotient} or \textit{generalized Rayleigh quotient} is defined as \cite{parlett1998symmetric}:
\begin{align}\label{equation_generalized_rayleigh_ritz_quotient}
\mathbb{R} \ni R(\b{A}, \b{B}; \b{x}) := \frac{\b{x}^\top \b{A}\, \b{x}}{\b{x}^\top \b{B}\, \b{x}},
\end{align}
where $\b{A}$ and $\b{B}$ are symmetric matrices and $\b{x}$ is a non-zero vector:
\begin{align}
\b{A} = \b{A}^\top, ~~ \b{B} = \b{B}^\top, ~~ \b{x} \neq \b{0}.
\end{align}
If the symmetric $\b{B}$ is positive definite:
\begin{align}
\b{B} \succ 0,
\end{align}
it has a Cholesky decomposition:
\begin{align}\label{equation_generalized_rayleigh_ritz_quotient_B_Cholesky}
\b{B} = \b{C} \b{C}^\top,
\end{align}
where $\b{C}$ is a lower triangular matrix. 
In case $\b{B} \succ 0$, the generalized Rayleigh-Ritz quotient can be converted to a Rayleigh-Ritz quotient:
\begin{align}
R(\b{A}, \b{B}; \b{x}) = R(\b{D}, \b{C}^\top \b{x}),
\end{align}
where:
\begin{align}\label{equation_generalized_rayleigh_ritz_quotient_D}
\b{D} := \b{C}^{-1} \b{A} \b{C}^{-\top}.
\end{align}
The proof is:
\begin{align*}
\text{RHS} &= R(\b{D}, \b{C}^\top \b{x}) \overset{(\ref{equation_rayleigh_ritz_quotient})}{=} \frac{(\b{C}^\top \b{x})^\top \b{D}\, (\b{C}^\top \b{x})}{(\b{C}^\top \b{x})^\top (\b{C}^\top \b{x})} \\
&\overset{(\ref{equation_generalized_rayleigh_ritz_quotient_D})}{=} \frac{\b{x}^\top \b{C} \b{C}^{-1} \b{A} (\b{C} \b{C}^{-1})^\top \b{x}}{\b{x}^\top (\b{C} \b{C}^\top) \b{x}} \overset{(a)}{=} \frac{\b{x}^\top \b{A}\, \b{x}}{\b{x}^\top \b{B}\, \b{x}} \\
&\overset{(\ref{equation_generalized_rayleigh_ritz_quotient})}{=} R(\b{A}, \b{B}; \b{x}) = \text{LHS}, ~~~~~ \text{Q.E.D.},
\end{align*}
where RHS and LHS are short for right and left hand sides and $(a)$ is because of Eq. (\ref{equation_generalized_rayleigh_ritz_quotient_B_Cholesky}) and $\b{C}\b{C}^{-1} = \b{I}$ because $\b{C}$ is a square matrix.

Similarly, one of the properties of the generalized Rayleigh-Ritz quotient is:
\begin{align}\label{equation_generalized_rayleigh_ritz_quotient_scaling}
R(\b{A}, \b{B}; c\b{x}) = R(\b{A}, \b{B}; \b{x}),
\end{align}
where $c$ is a scalar. The proof is that:
\begin{align*}
R(\b{A}, \b{B}; c\b{x}) &= \frac{(c\b{x})^\top \b{A}\, c\b{x}}{(c\b{x})^\top \b{B}\, c\b{x}} \overset{(a)}{=} \frac{c\b{x}^\top \b{A}\, c\b{x}}{c\b{x}^\top \b{B}\, c\b{x}} \\
&\overset{(b)}{=} \frac{c^2}{c^2} \times \frac{\b{x}^\top \b{A}\, \b{x}}{\b{x}^\top \b{B}\, \b{x}} \overset{(\ref{equation_generalized_rayleigh_ritz_quotient})}{=} R(\b{A}, \b{B}; \b{x}),
\end{align*}
where $(a)$ and $(b)$ are because $c$ is a scalar.

Because of the Eq. (\ref{equation_generalized_rayleigh_ritz_quotient_scaling}), the optimization of the generalized Rayleigh-Ritz quotient has an equivalent:
\begin{align}
&\underset{\b{x}}{\text{minimize/maximize}} ~~ R(\b{A}, \b{B}; \b{x}) \equiv \nonumber \\
&\begin{aligned}\label{equation_generalized_rayleigh_ritz_quotient_optimization}
& \underset{\b{x}}{\text{minimize/maximize}}
& & \b{x}^\top \b{A}\, \b{x} \\
& \text{subject to}
& & \b{x}^\top \b{B}\, \b{x} = 1,
\end{aligned}
\end{align}
for a similar reason that we provided for the Rayleigh-Ritz quotient. the constraint can be equal to any constant because in the derivative of Lagrangian, the constant will be dropped.

\subsection{Rayleigh-Ritz Quotient for Matrix}

The variable can be a matrix $\b{X}$ rather than vector $\b{x}$.
The Rayleigh-Ritz quotient and generalized Rayleigh-Ritz quotient for matrix $\b{X}$ are:
\begin{align}
&\mathbb{R} \ni R(\b{A}, \b{X}) := \frac{\textbf{tr}(\b{X}^\top \b{A}\, \b{X})}{\textbf{tr}(\b{X}^\top \b{X})}, \\
&\mathbb{R} \ni R(\b{A}, \b{B}; \b{X}) := \frac{\textbf{tr}(\b{X}^\top \b{A}\, \b{X})}{\textbf{tr}(\b{X}^\top \b{B} \b{X})},
\end{align}
respectively, where $\textbf{tr}(.)$ denotes the trace of matrix. 

For matrix variable, Eqs. (\ref{equation_rayleigh_ritz_quotient_optimization}) and (\ref{equation_generalized_rayleigh_ritz_quotient_optimization}) are \textit{approximately}, and not exactly, true. 
These equations are \textit{exact} for vector variable, as were proved. 
For matrix variable, we approximately have:
\begin{align}
&\underset{\b{x}}{\text{minimize/maximize}} ~~ \frac{\textbf{tr}(\b{X}^\top \b{A}\, \b{X})}{\textbf{tr}(\b{X}^\top \b{X})} \cong \nonumber \\
&\begin{aligned}\label{equation_rayleigh_ritz_quotient_optimization_matrix}
& \underset{\b{x}}{\text{minimize/maximize}}
& & \textbf{tr}(\b{X}^\top \b{A}\, \b{X}) \\
& \text{subject to}
& & \textbf{tr}(\b{X}^\top \b{X}) = 1,
\end{aligned}
\end{align}
and:
\begin{align}
&\underset{\b{x}}{\text{minimize/maximize}} ~~ \frac{\textbf{tr}(\b{X}^\top \b{A}\, \b{X})}{\textbf{tr}(\b{X}^\top \b{B} \b{X})} \cong \nonumber \\
&\begin{aligned}\label{equation_generalized_rayleigh_ritz_quotient_optimization_matrix}
& \underset{\b{x}}{\text{minimize/maximize}}
& & \textbf{tr}(\b{X}^\top \b{A}\, \b{X}) \\
& \text{subject to}
& & \textbf{tr}(\b{X}^\top \b{B} \b{X}) = 1.
\end{aligned}
\end{align}

\bibliography{References}
\bibliographystyle{icml2016}

\end{document}